\documentclass[conference]{IEEEtran}
\IEEEoverridecommandlockouts
\usepackage{cite}
\usepackage{amsmath,amssymb,amsfonts}
\usepackage{graphicx}
\usepackage[caption=false]{subfig}
\usepackage[utf8]{inputenc}
\usepackage[T1]{fontenc}
\usepackage{bbm}
\DeclareMathOperator*{\argmax}{arg\,max}
\DeclareMathOperator*{\argmin}{arg\,min}
\DeclareMathOperator*{\MSE}{MSE}
\DeclareMathOperator*{\Bias}{Bias}
\usepackage{booktabs}
\usepackage{xcolor}
\usepackage{tikz}
\usepackage{float}
\usepackage{textcomp}
\usepackage{tikz}
\usepackage{textcomp}

\newcommand\copyrighttext{%
  \footnotesize \textcopyright 2020 IEEE. Personal use of this material is permitted. Permission from IEEE must be obtained for all other uses, in any current or future media, including reprinting/republishing this material for advertising or promotional purposes, creating new collective works, for resale or redistribution to servers or lists, or reuse of any copyrighted component of this work in other works.}
\newcommand\copyrightnotice{%
\begin{tikzpicture}[remember picture,overlay]
\node[anchor=south,yshift=10pt] at (current page.south) {\fbox{\parbox{\dimexpr\textwidth-\fboxsep-\fboxrule\relax}{\copyrighttext}}};
\end{tikzpicture}%
}

\def\BibTeX{{\rm B\kern-.05em{\sc i\kern-.025em b}\kern-.08em
    T\kern-.1667em\lower.7ex\hbox{E}\kern-.125emX}}
\begin{document}

\title{\LARGE \bf Off-policy Learning for Remote Electrical Tilt Optimization}

\author{\IEEEauthorblockN{Filippo Vannella\IEEEauthorrefmark{1}\IEEEauthorrefmark{2},
Jaeseong Jeong\IEEEauthorrefmark{2}, Alexandre Proutiere\IEEEauthorrefmark{1}}
\IEEEauthorblockA{\IEEEauthorrefmark{1}KTH Royal Institute of Technology, Stockholm, Sweden \\
\IEEEauthorrefmark{2}Ericsson Research, Stockholm, Sweden\\
Email: vannella@kth.se, jaeseong.jeong@ericsson.com, alepro@kth.se}
}

\maketitle
\copyrightnotice

\thispagestyle{empty}
\pagestyle{empty}

\begin{abstract}
We address the problem of Remote Electrical Tilt (RET) optimization using off-policy Contextual Multi-Armed-Bandit (CMAB) techniques. The goal in RET optimization is to control the orientation of the vertical tilt angle of the antenna to optimize Key Performance Indicators (KPIs) representing the Quality of Service (QoS) perceived by the users in cellular networks. Learning an improved tilt update policy is hard. On the one hand, coming up with a new policy in an online manner in a real network requires exploring tilt updates that have never been used before, and is operationally too risky. On the other hand, devising this policy via simulations suffers from the simulation-to-reality gap. In this paper, we circumvent these issues by learning an improved policy in an offline manner using existing data collected on real networks. We formulate the problem of devising such a policy using the off-policy CMAB framework. We propose CMAB learning algorithms to extract optimal tilt update policies from the data. We train and evaluate these policies on real-world 4G Long Term Evolution (LTE) cellular network data. Our policies show consistent improvements over the rule-based logging policy used to collect the data.
\end{abstract}

\section{Introduction}
We focus on the automation of the Remote Electrical Tilt (RET) of Base Stations' (BSs) antennas in 4G Long Term Evolution (LTE) cellular networks. The RET problem consists in remotely controlling the vertical tilt angle of multiple antennas to optimize ad-hoc Key Performance Indicators (KPIs) such as coverage, capacity, interference, etc, that determine the network Quality of Service (QoS). RET automation can be framed in the context of Self-Optimizing Networks (SON), the network automation technology introduced by the $3^{rd}$ Generation Partnership Project (3GPP), and is an effective technique for handling interference and improving coverage and capacity in mobile networks \cite{Waldhauser11}.

Traditionally, the planning of antenna tilt angle in wireless mobile network has been left to expert knowledge, solving the problem through handcrafted rule-based algorithms. Conventional approaches to RET optimization consist of heuristic control strategies designed through domain knowledge, and mainly based on the optimization of utility metrics (see e.g. \cite{Eckhardt11, Partov15, Dandanov17}), or threshold-based policies employing Fuzzy Logic (FL) on representative network KPIs \cite{Saeed12,Buenestado16}. However, due to the growing sophistication of cellular networks, hand-crafted procedures for RET optimization are becoming increasingly more complex and time consuming, often leading to sub-optimal solutions and lack of adaptability. In fact, it is difficult to find an expert tilt tuning algorithm that can provide optimal radio propagation towards User Equipment (UE) while taking into account complex effects such as stochastic channel models, irregular antenna patterns, inter-cell interference, and highly variable user locations and demand. For these reasons, we need new approaches to RET optimization to improve network performance and reduce operational costs \cite{Khirallah11}.

Recently proposed methods for RET optimization are essentially data-driven learning approaches, mainly based on learning techniques used in Reinforcement Learning (RL) \cite{Guo13,Balevi19}, in Contextual Multi-Armed-Bandit (CMAB) or Multi-Armed Bandit (MAB) \cite{Cai10,Gulati14,Dhahri17}. Methods combining FL and RL techniques have also been investigated  \cite{Razavi10, Li12, Shaoshuai14}. There, the main idea is to use FL to encode a threshold-based discrete state-action space and to use it in learning algorithms based on $Q$-learning. In all these recent methods, an agent learns an optimal tilt update policy by directly interacting with the system and collecting feedback signals as a consequence of her actions. To learn an optimal policy, the agent needs to explore actions that have never been tested, which in turn, may lead the system to unsafe or low-performance states while learning. This issue is not tolerated in many real-world use cases and greatly limits the applicability of data-driven interactive learning algorithms \cite{Dulac19}. 

The whole body of literature concerning data-driven learning approaches to RET control, avoids this important practical problem by making use of a simulated environment to learn the new policy, i.e. by letting the agent interacting with the offline simulator instead of the real-world system. Unfortunately, policies learned in simulation often fail when deployed to the real world due to the inherent discrepancies between the physical system and the simulation model also known as \textit{simulation-to-reality gap}.

We take a different approach, based on offline off-policy learning \cite{Dudik11, Swaminathan15, Swaminathan15b}. Specifically, we aim at learning an optimal policy from offline data collected another policy, referred to as the {\it logging} policy. While operating the network, vast amount of data is collected and stored by telecommunication operators at little or no cost. These offline datasets represent a significant advantage for learning policies when compared to online approaches where the agent is required to learn in a trial and error fashion that inevitably degrades the network's performance during exploration phases. However, learning a new policy completely offline gives rise to new challenges that are not contemplated in the online setting. In particular, the dataset collected under the logging policy may have a strong bias towards actions that are very frequent under this policy. This issue is exacerbated by the inherent partial (often referred to as {\it bandit}) feedback available in the dataset (only feedback from actions executed by the logging policy are observed). 

We address these challenges by modelling the RET optimization problem as a CMAB problem and by exploiting two important techniques used in off-policy learning for CMAB problems: Inverse Propensity Scoring (IPS) and Direct Method (DM). We parametrize the learning policy as Artificial Neural Networks (ANNs) and devise off-policy learning algorithms based on IPS and DM. Our contributions are three-fold:

\begin{enumerate}
    \item We model the offline learning problem of an optimal RET policy as an off-policy CMAB problem.
    \item We devise and evaluate offline learning algorithms based on IPS or DM. Our experiments reveal that that the proposed algorithms outperform the currently deployed rule-based RET optimization strategy on real-world LTE network data. To the best of our knowledge, this is the first paper experimenting RET performance on real network data.
    \item Our analytical and experimental results further show that algorithms based on DM rather than IPS work better for RET off-policy learning where the KPI measurements include a significant exogenous noise component.
\end{enumerate}

\section{Background}
In this section we describe the mobile network system model considered for the RET problem, and provide an introduction to the CMAB framework.

\subsection{System model}
\label{sec:RET_model}
\begin{figure}[!h]
    \centering
     \includegraphics[width=\columnwidth]{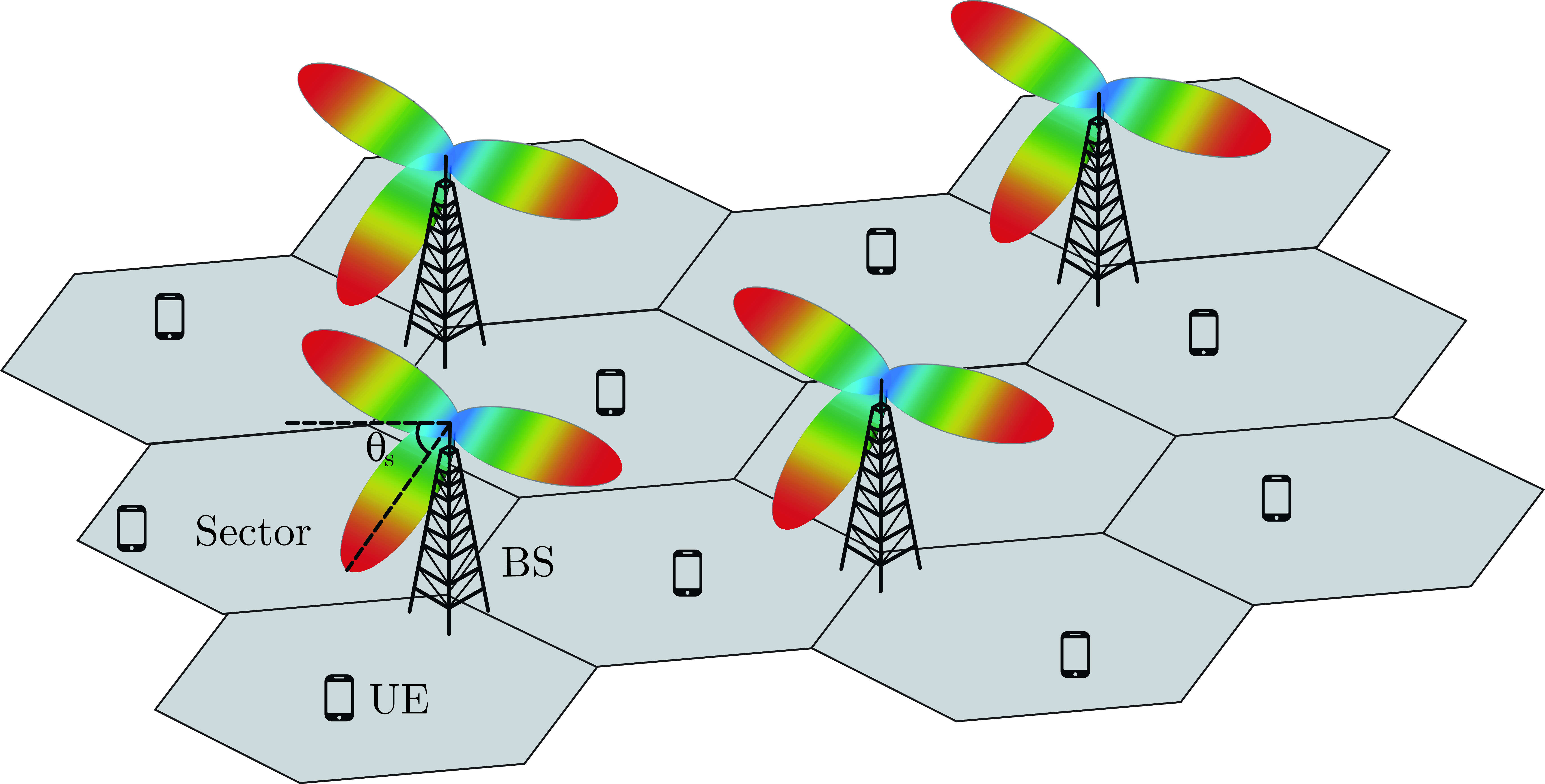}
     \caption{Abstract representation of the mobile network environment.}
     \label{fig:env}
\end{figure}

We consider a multi-cell and sectorized wireless mobile network covered by BSs. Each BS is endowed with three-sector antennas, covering three different sectors approximated by hexagonal shapes as in Fig. \ref{fig:env}.
\\ The network has $S$ antennas covering a given geographical area. The RET degree for sector $s$ is by denoted by $\theta_s$. It is defined as the inclination of the main lobe of the antenna radiation pattern with respect to its horizontal plane (see Fig. \ref{fig:env}). Inspired by previous works in the Coverage and Capacity Optimization (CCO) literature \cite{Dandanov17, Dhahri17, Li12}, we consider two KPIs for each sector $s$ and each period: coverage~$q_s \in [0,1]$ and capacity $c_s \in [0,1]$. The joint optimization of these two trading-off KPIs aims at maximizing the network capacity while ensuring that the targeted service areas remain covered. In this paper, we deem that $q_s$ and $c_s$ are risk-alarming KPIs, i.e., higher value indicates worse performance of the corresponding KPI. Both $q_s$ and $c_s$ are calculated by raw KPIs such as Reference Signal Receive Power (RSRP), Radio Resource Control (RRC) congestion rate, measured in each sector and also contains information about KPIs in neighboring sectors. The coverage $q_s$ mainly reflects the radio coverage range and edge performance of the sector $s$, while the capacity~$c_s$ considers the signal strength, the sector congestion and the interference from neighboring sectors. 

We consider a RET control policy, whose input consists of observed KPIs in a sector $s$, i.e., $(q_s,c_s),$ and output is a decision on whether to \textit{up-tilt}, \textit{down-tilt} or \textit{no-change}~$\theta_s$. Such RET control policy is executed per each sector $s$ periodically (e.g. every weekday, every hour, etc.). The KPIs $q_s$ and $c_s$ at each execution are computed by aggregating raw KPIs measured from the time of the previous execution.
The control strategy fine-tunes the antenna tilt on sector-by-sector basis and is global, meaning that the same policy (a mapping between the KPIs and the action) is executed at all $S$ sectors. 

\subsection{Contextual Multi-Armed-Bandit off-policy learning}
\label{sec:MAB}
The CMAB is a sequential decision making problem in which, at the beginning of each period, an agent observes a context $x$, and executes an action $a$; and at the end of the period, she observes a noisy loss $\delta$. The input $x \in \mathcal{X}$, is assumed to be drawn from an unknown probability distribution $x \sim p(\cdot)$. The action $a \in \mathcal{A}$ is sampled from a policy $a \sim \pi(\cdot|x)$, that maps contexts to probability distributions over actions. As a consequence for executing action $a$ given the context $x$, the agent experiences a noisy loss~$\delta \sim \Delta(\cdot|x,a)$. Given a (context, action) pair, the loss distribution $\Delta(\cdot |x,a)$ has a mean $\bar{\delta}(x, a) \triangleq \mathbb{E}_{\Delta(\cdot|x,a)}[\delta|x,a]$ and a variance  
$\sigma_{\delta}^{2}(x, a)  \triangleq \mathbb{V}_{\Delta (\cdot|x,a)}[\delta|x,a]$. 

In the off-policy learning CMAB setting, we leverage a dataset $\mathcal{D}_{\lambda} = \{(x_i,a_i,\delta_i)\}_{i = 1}^{N}$ collected under the logging policy $\lambda$, where $x_i\sim p(\cdot)$, $a_i \sim \lambda(\cdot|x_i)$, and $\delta_i \sim \Delta(\cdot |x_i,a_i)$\footnote{For notational convenience, we use $\delta_i$ and  $\delta(x_i,a_i)$ interchangeably.}. From this dataset, we wish to be able to evaluate the performance of a {\it target} policy $\pi$, and to learn an optimal policy $\pi^\star$ among a given class $\Pi$ of policies. The performance of a target policy $\pi$ is assessed via its {\it risk}, defined as:
\begin{equation}
\label{eq:noisyR}
\begin{aligned}
    R\left(\pi\right) &= \mathbb{E}_{x\sim p(\cdot)}\mathbb{E}_{a\sim \pi(\cdot|x)}\left[\bar{\delta}(x,a)\right].
\end{aligned}\end{equation}
For simplicity, in the remainder of the paper, we use the following notation: $\mathbb{E}_{\pi}\left[\cdot\right] \triangleq \mathbb{E}_{x\sim p(\cdot)}\mathbb{E}_{a\sim\pi\left(\cdot\middle| x\right)}\mathbb{E}_{\delta\sim \Delta(\cdot|x,a)}\left[\cdot\right]$. Given a class of policies $\Pi$, an optimal target policy $\pi^\star$ is such that: 
\begin{equation}
    \pi^\star \in \argmin_{\pi\in\Pi} R\left(\pi\right).
\label{eq:obj}
\end{equation}
The distributions $p(\cdot)$ and $\Delta(\cdot|x,a)$ are unknown, and so the risk of a policy $\pi$ is impossible to compute exactly. Instead, we need to estimate the risk using the dataset. The challenge is to design an accurate estimator $\hat R(\pi)$ of $R(\pi)$. Based on $\hat R(\pi)$, we can then select an approximately optimal policy as
\begin{equation}
    \pi^\star = \argmin_{\pi\in\Pi} \hat R\left(\pi\right).
\label{eq:hatobj}
\end{equation}


\section{Off-policy learning for RET optimization}
\label{sec:off_policy_ret_opt}
In this section, we formulate the RET control problem using the CMAB off-policy learning framework, and describe learning algorithms based on IPS and DM risk estimators.

\subsection{CMAB formulation of the offline RET problem}
\label{sec:RET_off_policy}
Starting from the framework described in Section \ref{sec:MAB}, we proceed to define context, action, and loss in the case of the RET optimization problem. We define the context as $x = (q_s,c_s) \in [0,1]^2$, that is, we consider the aggregated risk-alarming KPIs at sector $s$ for coverage $c_s$ and capacity $q_s$ measured in the time period before action execution, introduced in Section \ref{sec:RET_model}. The discrete action space is $\mathcal{A}=\{-\varepsilon, 0,\varepsilon\}$, where $\varepsilon$ is the amount of tilt degree change. The loss metric $\delta$ is an hand-crafted indicator of the sector performance degradation as a consequence of the tilt change action. Denoting $(c^\prime_s,q^\prime_s)$ the risk-alarming KPIs after the execution of the control action $a$, one example definition of the loss could be $\delta = \max(c^\prime_s,q^\prime_s) - \max(c_s,q_s)$. 

The logging policy $\lambda(\cdot|x)$ is a rule-based policy controlling~$\theta_s$ in sector $s$. The execution of $\lambda$ is on a sector-by-sector basis, and the same rule is applied in all sectors. $\lambda$ is used for generating $N$ data observations from sectors of 4G mobile networks, that are collected in a dataset denoted by $\mathcal{D}_{\lambda} = \{(x_i,a_i,\delta_i)\}_{i = 1}^{N}$. We estimate the probability distribution defining the logging policy $\lambda$ from $\mathcal{D}_{\lambda}$ using a logistic regression model.
With $\mathcal{D}_{\lambda},$ our goal is to learn an optimal policy $\pi^\star$ based on the risk minimization objective in \eqref{eq:hatobj}.

As mentioned already, the main challenge lies in the estimation of the risk of a given target policy from the dataset~$\mathcal{D}_{\lambda}.$ First, the observed loss $\delta_i$ in $\mathcal{D}_{\lambda}$ includes a significant amount of noise (with variance $\sigma_\delta^2$) due to exogenous factors in the network such as random traffic demand, human mobility, etc. Second, samples in $\mathcal{D}_{\lambda}$ exhibit a very unbalanced action distribution, because the deployed logging policy $\lambda$ takes no-change action (i.e., $a=0$) much more often than up-tilt or down-tilt actions in order to keep a conservative network operation. In the following subsections, we discuss how the risk estimator based on the IPS and the DM addresses these challenges.

\subsection{Off-policy risk estimation methods}
We focus on two types of off-policy estimator $\hat{R}(\pi)$ and analyze their bias and variance, resulting from the noise in the measured loss and unbalanced action probabilities in the logging policy $\lambda$.

\subsubsection{IPS risk estimator} This estimator of $R(\pi)$ consists in re-weighting the sampled losses by the inverse of the  probability of the observed action under the logging policy \cite{Swaminathan15}. Specifically when deriving $\hat{R}(\pi)$, a sample $(x,a,\delta)$ is weighted by $w(x,a) = \frac{\pi(a|x)}{\lambda(a|x)}$. This weight is referred to as the IPS weight, or likelihood ratio, and it is used to correct for the distribution mismatch between $\pi$ and $\lambda$. The IPS weight is well defined only if $\pi$ is absolutely continuous w.r.t. $\lambda$, i.e. for all context-action pairs $(x,a) \in \mathcal{X}\times \mathcal{A}$, we have $\pi(a|x) > 0 \Rightarrow  \lambda (a|x) > 0$. The IPS risk estimator is given by:
\begin{equation}
    \hat{R}_{\mathrm{IPS}}(\pi) = \frac{1}{N}\sum_{i = 1}^N w(x_i,a_i) \cdot \delta(x_i,a_i).
    \label{eq:IS}
\end{equation}
In case of independent and identically distributed (\textit{i.i.d}) context realizations, we have the following results for bias and variance \cite{Dudik11}: 
\begin{equation}
    \label{eq:IPS_statistics}
    \begin{aligned}
    \Bias\left[\hat{R}_{\mathrm{IPS}}(\pi)\right] & = \left|\mathbb{E}_\lambda[\hat{R}_{\mathrm{IPS}}(\pi)] - R(\pi)\right| = 0, 
    \\
    \mathbb{V}_{\lambda}\left[\hat{R}_{\mathrm{IPS}}(\pi)\right] &=  \frac{\mathbb{E}_{\lambda} \left[\sigma_{\delta}^{2}(x, a)w^{2}(x, a)\right]}{N} \\& + \frac{\mathbb{V}_{\lambda}\left[\bar{\delta}(x,a) w(x, a)\right]}{N}.
      \end{aligned}
\end{equation}
Thus, the IPS estimator is unbiased, but it may suffer from high variance, that gets worse as the discrepancy between the logging policy and the target policy $w(x,a)$ or the loss noise variance $\sigma_\delta^2$ increases.

\subsubsection{DM risk estimator} This estimator consists in estimating the loss function $\bar{\delta}$ from the data and in using this estimated loss to assess the performance of the target policy \cite{Dudik11}. More precisely, we seek for an estimate $\hat{\delta}:\mathcal{X}\times \mathcal{A} \rightarrow \mathbb{R}$ of the expected loss $\bar{\delta}$ within a given class of function ${\cal F}$, e.g. parameterized by an ANN. Given a class ${\cal F}$ of functions, the objective is to minimize the (empirical) MSE between noisy and estimated loss:
\begin{equation}
    \begin{aligned}
       \hat{\delta}^\star &= \argmin_{\hat{\delta} \in \mathcal{F}} \widehat{\MSE}(\hat{\delta}) \\ & = \argmin_{\hat{\delta} \in \mathcal{F}} \frac{1}{N}\sum_{i=1}^{N}{[\delta(x_i,a_i)}-{\hat{\delta}}(x_i,a_i)]^2.
       \label{eq:MSE}
    \end{aligned}
\end{equation}

Now under the DM method, the risk estimator is:
\begin{equation}
    \hat{R}_{\mathrm{DM}}(\pi)=\frac{1}{N} \sum_{i= 1}^N \sum_{a\in\mathcal{A}} \pi(a|x_i)\hat{\delta}^\star(x_i,a).
    \label{eq:DM}
\end{equation}
The quality of the DM risk estimator $\hat{R}_{DM}(\pi)$ heavily relies on that of the loss estimation procedure. If $\hat{\delta}(x,a)$ forms a good approximation of the true expected loss $\bar{\delta}(x,a)$, then the DM estimator will also be accurate. 
Regarding the bias-variance properties of this estimator, assuming again \textit{i.i.d.} context realizations, we have (see e.g. \cite{Dudik11}):
\begin{equation}
    \label{eq:DM_statistics}
    \begin{aligned}
     \Bias\left[\hat{R}_{\mathrm{DM}}(\pi)\right]& =  \left|\mathbb{E}_{\lambda}\left[\hat{R}_{\mathrm{DM}}(\pi)\right]-R(\pi)\right|\\ & =\left|\mathbb{E}_{\pi}[\hat{\delta}^\star(x,a) -\bar{\delta}(x,a)]\right|,
    \\ 
    \mathbb{V}_{\lambda}\left[\hat{R}_{\mathrm{DM}}(\pi)\right] & = \frac{1}{N} \mathbb{V}_{\pi}\left[\hat{\delta}^\star(x,a)\right]. 
   \end{aligned}
   \end{equation}
Thus, we observe that the DM risk estimator is in general biased, depending on the bias of $\hat{\delta}^\star(x,a)$. In general, DM methods encounter problems when the loss model do not represent well the loss in areas of the context-action space that are important for the target policy. However the variance of DM estimators can be consistently lower than that of the IPS estimator (it scales as $\frac{1}{N}$ times the variance of $\hat{\delta}^\star$). 

\subsection{Off-policy learning methods}
We describe how to learn an optimal policy $\pi^\star$ (with respect to the objective \eqref{eq:hatobj}) when using the IPS or the  DM risk estimator. We model both the target policy and the loss through ANNs. This choice is motivated by the lack of a-priori information about the structure of the optimal policy or loss function; in such a case, the use of ANNs, a class of universal function approximators, is instrumental. 

\subsubsection{IPS off-policy learning}
we denote the ANN parametrizing the target policy as $\pi(a|x;w) \triangleq\pi_{w}(a|x)$  having weight vector $w \in \mathbb{R}^p$. The resulting learning objective for the IPS estimator is 
\begin{equation}
\begin{aligned}
    w^\star &= \argmin_{w \in \mathbb{R}^p} \hat{R}_{IPS}(\pi_{w}) \\&= \argmin_{w \in \mathbb{R}^p} \frac{1}{N}\sum_{i = 1}^N \frac{\pi_{w}(a_i|x_i)}{\lambda(a_i|x_i)}\delta(x_i,a_i)
    \label{eq:RIS_obj}
    \end{aligned}
\end{equation}

After solving \eqref{eq:RIS_obj}, we consider the greedy deterministic policy based on IPS estimator as $ \hat{\pi}_{\mathrm{IPS}}^\star(a|x) = \mathbbm{1}\{a = \argmax_{b\in\mathcal{A}} \pi_{w^\star}(x, b)\}$.

\subsubsection{DM off-policy learning}
we denote the ANN parametrizing the loss as $\hat{\delta}_v(x,a) \triangleq \hat{\delta}(x,a;v)$, having weight vector $v \in \mathbb{R}^q$. The training objective for loss estimation is
\begin{equation}
     \label{eq:MSE_min}
     \begin{aligned} 
     {v^\star} &= \argmin_{v\in\mathbb{R}^q}\widehat{\MSE}(\hat{\delta}_v) \\ &=\argmin_{v\in\mathbb{R}^q} \frac{1}{N}\sum_{i=1}^{N}{[\delta\left(x_i,a_i\right)}-{\hat{\delta}_{v}}\left(x_i,a_i\right)]^2
     \end{aligned}
\end{equation}

Once $\hat{\delta}_v^\star$ is estimated, the deterministic greedy policy derived for the DM estimator is $\hat{\pi}^\star_{DM}(a|x)=\mathbbm{1}\{a = \argmin_{\bar{a}\in\mathcal{A}} \hat{\delta}_{v^\star}(x, \bar{a})\}$.

\section{Experiments and Results}
\label{sec:exp_setup}
In this section, we present the experimental setup and we empirically evaluate our off-policy learning algorithms based on either the IPS or the DM estimator.
\subsection{Experimental setup}
The dataset $\mathcal{D}_{\lambda}$ contains $N = 309435$ data points collected from real-world 4G networks where $\lambda$ has been executed for RET control. We split the data into training ($70\%$) and testing ($30\%$) datasets, denoted by $\mathcal{D}_{\lambda}^{\text{train}}$ and $\mathcal{D}_{\lambda}^{\text{test}} $ respectively, such that $|\mathcal{D}_{\lambda}^{\text{train}}| = N_{\text{train}} = 216605 $, and $|\mathcal{D}_{\lambda}^{\text{test}}| = N_{\text{test}} = 92830$. We validate the experiments by executing $K = 5$ random and independent splits and reporting mean and standard deviation of the test performance results over the $K$ independent splits. 

In order to solve the optimization problems in \eqref{eq:MSE_min} and \eqref{eq:RIS_obj}, we initialize the weights for the policy model $w \in \mathbb{R}^p$, and the loss model $v\in\mathbb{R}^q$ randomly, and apply mini-batch Adam optimizer 
with a batch size of $N_v^{\text{batch}} = N_w^{\text{batch}} = \frac{N_{\text{train}}}{100}$ and learning rates $\alpha_v = 0.001$ and $\alpha_w = 0.0005$. 
These values have been selected by executing grid hyper-parameters search for learning rate and batch size based on $\alpha = [0.0001,0.0005,0.001,0.005,0.01]$ and $N^{batch} =\left [\frac{N_{\text{train}}}{10}, \frac{N_{\text{train}}}{100} ,\frac{N_{\text{train}}}{1000}\right]$ for both $\delta_v$ and  $\pi_w$. 

The performance metric considered for the policy evaluation is the test loss based on IPS risk estimator:
\begin{displaymath}
    \label{eq:test_loss}
    \mathcal{L}(\hat{\pi}) = \frac{1}{\sum_{i = 1}^{N_\text{test}}\mathbbm{1}\{\hat{\pi}(a_i|x_i) = a_i\}}\sum_{i = 1}^{N_\text{test}} \frac{\delta_i \mathbbm{1}\{\hat{\pi}(a_i|x_i) = a_i\}}{\lambda(a_i|x_i)}.
\end{displaymath}

Recall that most of the actions executed by the logging policy $\lambda$ are no-change (i.e., $a=0$), because the deployed $\lambda$ aims at a conservative network operation. This makes the average loss of the logging policy $\lambda$ dominated by zero (i.e., performance does not vary because of the unchanged tilt degree). To address this issue, we also run the same evaluation on a down-sampled test dataset, where some of samples with action $a=0$ are removed. The down-sampled dataset is built so that the three possible actions are equally represented. Using this down-sampled dataset allows us to compare the loss with the proposed algorithms, accounting for the cases where $\lambda$ takes up-tilt or down-tilt actions.

\subsection{Results and discussion}
\begin{figure}[htp]
\begin{center}
\subfloat[Test loss for complete data.]{%
  \includegraphics[clip,width=0.7\columnwidth]{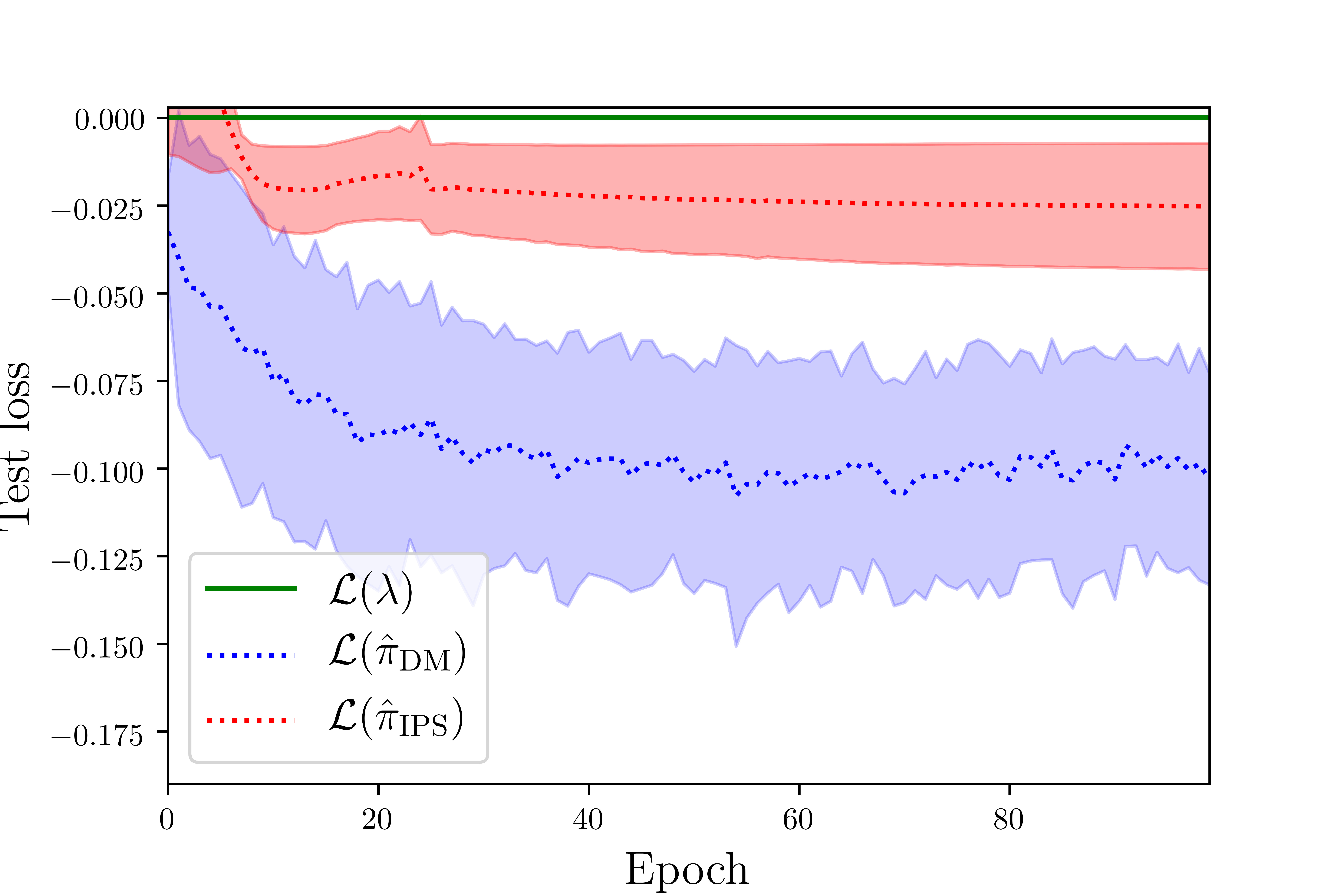}%
}

\subfloat[Test loss for down-sampled data.]{
  \includegraphics[clip,width=0.7\columnwidth]{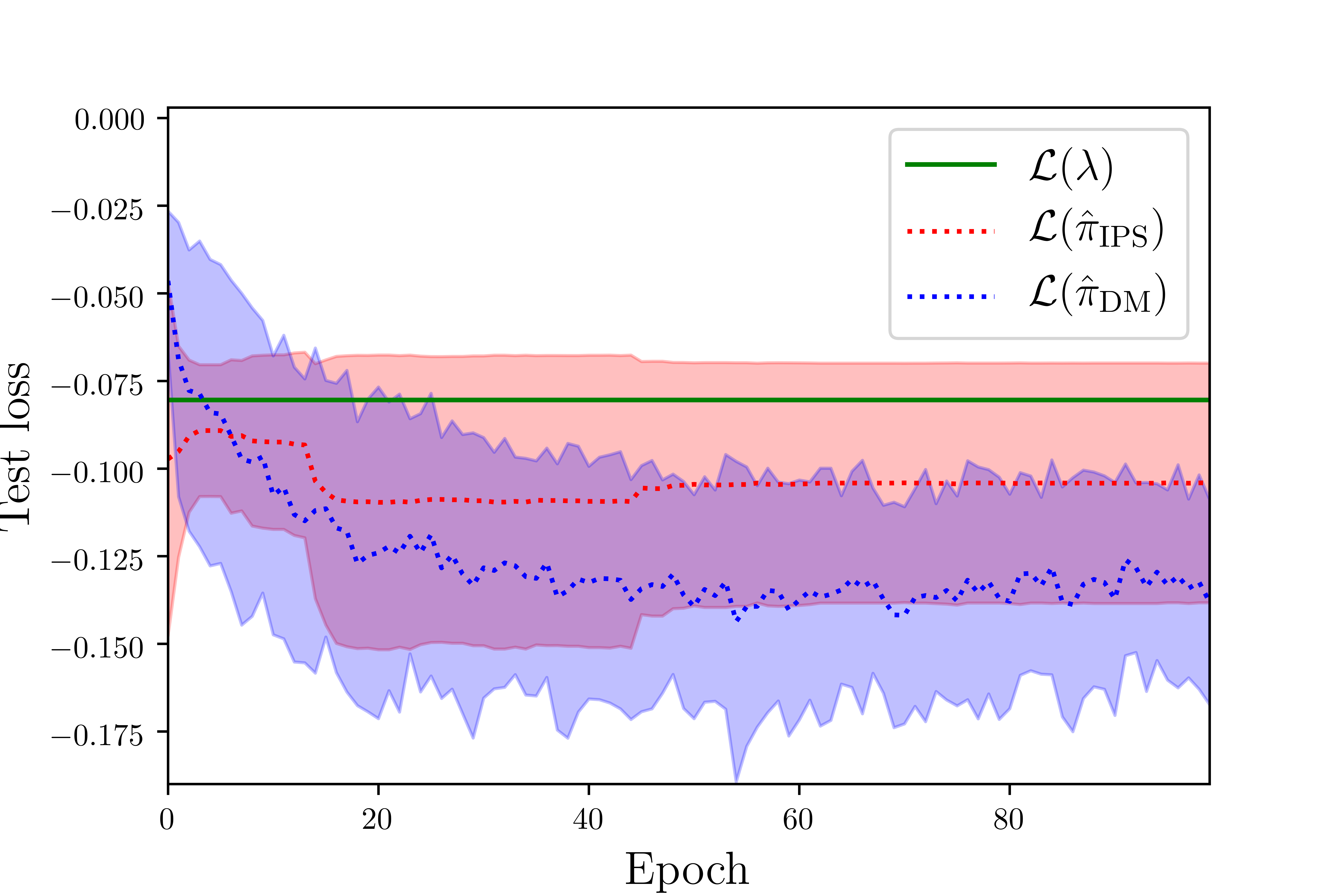}%
}
\caption{Test loss results.}
\label{fig:results}
\end{center}
\end{figure}

We test the performance of our algorithms as defined in \eqref{eq:test_loss} on  $\mathcal{D}_{\lambda}^{\text{test}}$ and its down-sampled counterpart. In Fig. \ref{fig:results}, the test loss is shown as a function of the {\it epoch}, representing the duration of the training procedure (one epoch corresponds to going through all samples of the training data once). Table \ref{tab:test} gives the test loss after 100 epochs for the DM and IPS policies.

\begin{figure}[h!]
    \centering
     \includegraphics[width=\columnwidth]{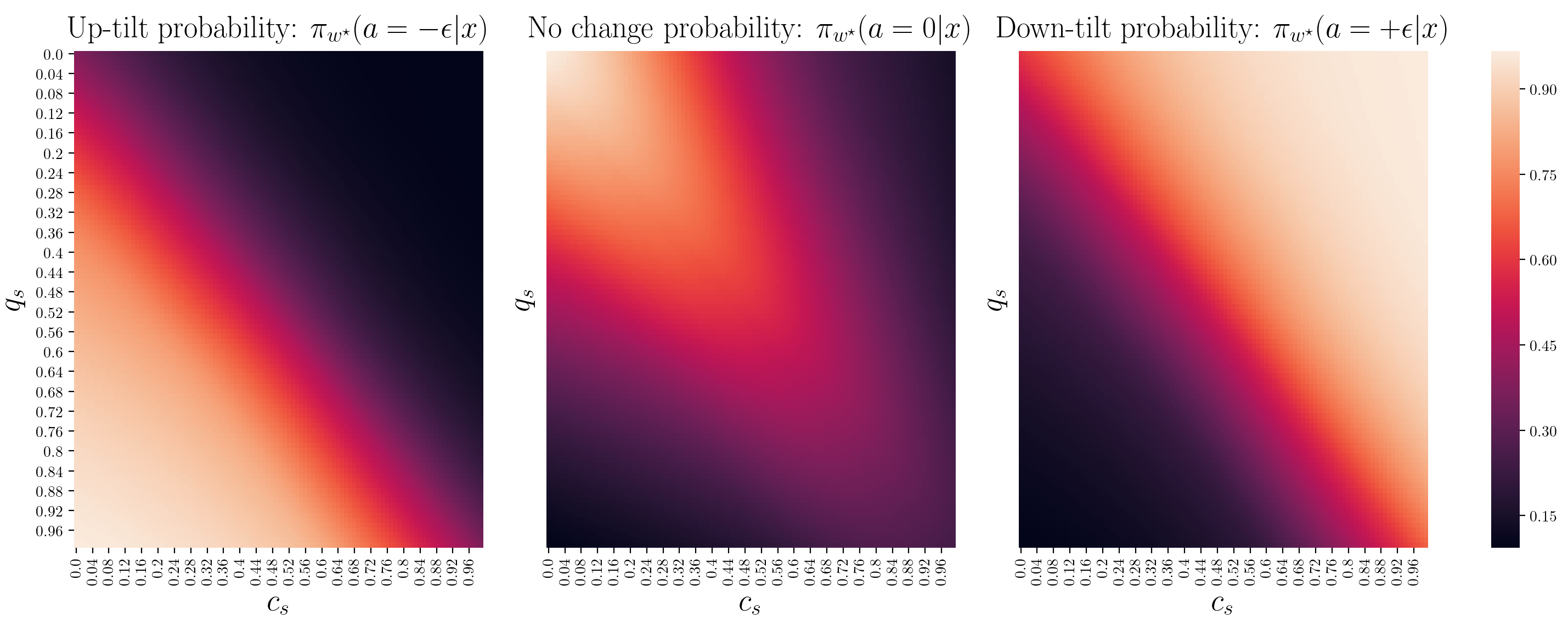}
     \caption{Heatmap for action probabilities of optimal IPS policy  $\pi_{w^\star}$}
     \label{fig:results_heatmap}
\end{figure}

From the experiments, we observe that the policy learnt by our off-policy training algorithms based on both IPS and DM outperforms the logging rule-based policy. We also observe that the DM learning algorithm produces a policy with higher performance than that obtained under the IPS learning algorithm. This confirms the DM and IPS bias-variance properties presented in Section \ref{sec:off_policy_ret_opt}: the DM estimator has a lower variance than the IPS estimator, which in turn results in a better policy especially in presence of significant noise in the loss. Indeed, our real-world network data exhibit a very significant level of noise.  
\begin{table}[h!]
\caption{Test loss results at convergence}
\label{tab:test}
\vskip 0.15in
\begin{center}
\begin{small}
\begin{sc}
\begin{tabular}{ccc}
\toprule
Estimator & Complete data & Down-sampled data\\
\midrule
$\mathcal{L}(\lambda)$ & $-0.00012 \pm 0.00003 $  & $-0.07924 \pm 0.00656$  \\
$\mathcal{L}( \hat{\pi}^\star_{\mathrm{IPS}})$  & $-0.02516 \pm 0.01797$ & $-0.10409 \pm 0.03418 $  \\
$\mathcal{L}( \hat{\pi}^\star_{\mathrm{DM}})$ & $-0.10307\pm 0.03015$ & $-0.13808 \pm 0.02927$  \\
\bottomrule
\end{tabular}
\end{sc}
\end{small}
\label{tab:data}
\end{center}
\vskip -0.1in
\end{table}
\\ 
In Fig. \ref{fig:results_heatmap}, we present the heatmap of the action probabilities of the learned IPS policy. 
\\ 
From Fig. \ref{fig:results_heatmap}, we observe that $\pi_{w^\star}$ puts an emphasis on the coverage KPI, favouring up-tilt actions for low values of coverage alarm, and down-tilt for high-values of coverage alarm, almost independently from the capacity alarm KPI. This may be due to the fact that the logging policy assigns higher weight to $c_s$, aiming at ensuring a minimum coverage rather than optimizing capacity. Finally, we observe that as expected the no-change action is executed in the subset of the context space where both $c_s$ and $q_s$ have low values.

\section{Conclusion}
In this paper, we have introduced a simple but effective data-driven method for RET optimization. We learn a policy using methods for offline off-policy learning in CMAB. Extensive experimental results on a real-world 4G LTE mobile network dataset have demonstrated the empirical effectiveness of the proposed methodology: the IPS and DM policies we learnt outperform the rule-based logging policy. As opposed to simulation-based RL techniques, the proposed off-policy method avoids simulators modelling errors by training a policy completely offline on data obtained directly from the network providing an higher degree of safety and reliability. Future directions include the design of learning objectives based on off-policy estimators that handles differently the bias-variance trade-off (e.g. Self-Normalized IPS (SNIPS) \cite{Swaminathan15b}, Doubly Robust (DR) \cite{Dudik11} estimators).

\section{Acknowledgements}
This work was partially supported by the Wallenberg AI, Autonomous Systems and Software Program (WASP) funded by the Knut and Alice Wallenberg Foundation.
\addtolength{\textheight}{-12cm}   
\bibliography{biblio.bib}

\end{document}